\pdfoutput=1
\documentclass[11pt]{article}
\usepackage[preprint]{acl}

\usepackage{times}
\usepackage{latexsym}
\usepackage[T1]{fontenc}
\usepackage[utf8]{inputenc}
\usepackage{microtype}
\usepackage{inconsolata}
\usepackage{graphicx}
\usepackage{amsmath}
\usepackage{url}
\usepackage{listings}
\usepackage{color}
\usepackage{hyperref}
\usepackage{caption}
\usepackage{booktabs}
\usepackage{tcolorbox}
\usepackage{multirow}

\title{Retrieval-Enhanced Few-Shot Prompting for Speech Event Extraction}

\author{Máté Gedeon \\
  Budapest University of Technology and Economics \\
  Budapest, Hungary \\
  {\tt gedeonm01@gmail.com} \\}

\date{}

\begin{document}
\maketitle
\begin{abstract}
Speech Event Extraction (SpeechEE) is a challenging task that lies at the intersection of Automatic Speech Recognition (ASR) and Natural Language Processing (NLP), requiring the identification of structured event information from spoken language. In this work, we present a modular, pipeline-based SpeechEE framework that integrates high-performance ASR with semantic search-enhanced prompting of Large Language Models (LLMs). Our system first classifies speech segments likely to contain events using a hybrid filtering mechanism including rule-based, BERT-based, and LLM-based models. It then employs few-shot LLM prompting, dynamically enriched via semantic similarity retrieval, to identify event triggers and extract corresponding arguments. We evaluate the pipeline using multiple LLMs—Llama3-8B, GPT-4o-mini, and o1-mini—highlighting significant performance gains with o1-mini, which achieves 63.3\% F1 on trigger classification and 27.8\% F1 on argument classification, outperforming prior benchmarks. Our results demonstrate that pipeline approaches, when empowered by retrieval-augmented LLMs, can rival or exceed end-to-end systems while maintaining interpretability and modularity. This work provides practical insights into LLM-driven event extraction and opens pathways for future hybrid models combining textual and acoustic features.
\end{abstract}

\section{Introduction}

Information extraction aims at automatically identifying structured information, such as entities and their relations, from unstructured data \cite{bikel1997, fei2023}. A task in this domain is Event Extraction (EE) \cite{chen2015} searching for answers to questions like \textit{what happened}, \textit{who was involved}, and \textit{where did it take place}. The source data can be text \cite{yang2016}, but even images \cite{li2020} or videos \cite{chen2021}. \emph{Speech Event Extraction} (SpeechEE) \cite{kang2024} extends textual EE, with the purpose of identifying structured event information directly from the input of the spoken language. This task occupies a unique intersection between Automatic Speech Recognition (ASR) and Natural Language Processing (NLP), requiring not only accurate transcription but also the detection of event types, triggers, and arguments from possibly noisy spoken content.

Existing SpeechEE approaches can be broadly categorized into the methodologies \emph{pipeline-based} and \emph{end-to-end}. Pipeline-based architectures typically employ an ASR module to transcribe speech, followed by text-based event extraction using NLP techniques \cite{cao2022}. These systems offer modularity and transparency, allowing separate optimization and analysis of ASR and extraction components. However, they are susceptible to cascading errors, where transcription inaccuracies can significantly impair downstream event extraction performance. Conversely, end-to-end approaches aim to bypass intermediate text by learning to map raw audio directly to structured outputs \cite{wang2024}. While promising in reducing error propagation and potentially more efficient, these models often demand large-scale annotated audio-event datasets and are less interpretable, acting as opaque "black boxes" in many cases.

Recent progress in Large Language Models (LLMs) such as GPT-4 \cite{gpt4} has opened new possibilities in pipeline-based architectures by enabling powerful few-shot and zero-shot learning capabilities. LLMs exhibit remarkable proficiency in extracting structured knowledge from unstructured text with minimal task-specific supervision \cite{zhang2022}. When combined with state-of-the-art ASR systems, these models can form the backbone of robust SpeechEE systems that generalize well across domains and require minimal adaptation.

In this work, we present a pipeline-based SpeechEE framework that leverages semantic search-enhanced few-shot prompting with LLMs. Our system dynamically retrieves relevant examples from a support set and incorporates them into prompts to guide event extraction. Additionally, we introduce a classification mechanism to identify utterances likely to contain events, reducing false positives and improving extraction precision.

Our main contributions are as follows:
\begin{itemize}
    \item We propose a multi-stage SpeechEE pipeline combining high-performance ASR with semantic search-enhanced prompting for event extraction using LLMs.
    \item We introduce a generalizable few-shot learning strategy based on semantic similarity, applicable to various text-related information extraction tasks.
    \item We develop a speech segment classification module that selectively filters utterances likely to contain events.
    \item We provide a detailed comparison of several configurations, offering practical insights for SpeechEE deployment.
\end{itemize}

The remainder of the paper is organized as follows. Section 2 reviews related work on SpeechEE and language models. Section 3 introduces the dataset. Section 4 describes the model architecture and pipeline components. Section 5 presents the experimental results and analysis. Section 6 outlines the results of the ablation study. Section 7 discusses key findings and limitations. Finally, Section 8 concludes the paper and suggests directions for future research.

\section{Related Work}
\subsection{Event Extraction from Text}
EE from textual data has been a long-standing task in information extraction, focusing on identifying event triggers and their semantic arguments. Early approaches were largely feature-based, relying on hand-engineered lexical, syntactic, and semantic features fed into statistical models such as maximum entropy classifiers, conditional random fields, or nearest-neighbor methods \cite{ahn2006} \cite{liao2010}. 

With the advent of deep learning, neural-based models became dominant. Convolutional Neural Networks (CNNs) enabled automatic feature learning from word embeddings, replacing manual feature engineering \cite{nguyen2015}. With the ability to handle long-range dependencies, recurrent architectures achieved state-of-the-art results on the ACE2005 dataset \cite{nguyen2016}.

Graph Convolutional Networks (GCNs) have also shown promise in the field by modeling syntactic structures directly from dependency trees. Unlike sequential models, GCNs exploit the syntactic proximity between triggers and arguments in a graph form, improving performance on long sentences with distant dependencies \cite{nguyen2018} \cite{liu2018}.

More recently, Transformer-based solutions \cite{vaswani2017} also emerged, with \cite{paolini2021} introducing a framework, achieving new state-of-the-art results on joint entity and relation extraction using a generative transformer.

A recent survey categorizes modern approaches into sequence-based, graph neural, knowledge-enhanced, and prompt-based methods, highlighting the dominance of pretrained language models in capturing contextual event semantics \cite{surveyee}.

\subsection{Speech Processing and ASR}
Automatic Speech Recognition (ASR) has witnessed remarkable progress in recent years, particularly with the introduction of transformer-based models. Whisper \cite{radford2022} represents a significant advancement in ASR, trained on a large and diverse dataset of 680,000 hours of multilingual and multitask supervised data. This approach has demonstrated robust performance across various domains and languages, making it suitable for real-world applications.

Similarly, Canary \cite{canary} offers an alternative approach to ASR that achieves competitive performance without relying on web-scale data. These models provide high-quality transcriptions that can serve as the foundation for downstream NLP tasks, including event extraction.

\subsection{Speech Event Extraction}
Speech Event Extraction (SpeechEE) is a relatively new research direction that aims to bridge ASR and event extraction. \cite{wang2024} introduced a novel benchmark for SpeechEE and proposed an end-to-end model for extracting events directly from audio. Their work highlights the challenges of extracting structured information from speech without relying on intermediate textual representations. Their end-to-end system demonstrated consistently superior performance compared to the employed pipeline-based baseline. They report that previous approaches to speech-based information extraction have largely relied on pipeline methods, where ASR is followed by text-based extraction. While these methods benefit from advances in both ASR and text-based event extraction, they often suffer from error propagation, where mistakes in transcription lead to extraction failures.

\subsection{Large Language Models and Few-Shot Learning}
Large Language Models (LLMs) have emerged as transformative tools in natural language processing (NLP), enabling significant progress across a wide range of tasks including text generation, summarization, machine translation, and information extraction. Proprietary models like GPT-4o \cite{gpt4o} and o1-mini \cite{openaio1} or open source models like Llama 3 \cite{llama3} exemplify the scale and versatility of these models. Their capacity to perform tasks with minimal or no explicit training—known as few-shot or zero-shot learning—has shifted the paradigm from model-specific fine-tuning to prompt-based task generalization.

Few-shot learning in LLMs typically involves crafting prompts that include a handful of task-specific examples, guiding the model to infer patterns and generalize to unseen inputs. This method leverages the latent knowledge encoded in the pretrained model, often yielding strong performance on tasks such as question answering, named entity recognition, and relation extraction \cite{gao2021} \cite{min2022}. Particularly for structured information extraction, few-shot prompting enables LLMs to identify and retrieve relevant spans of text even in the absence of large annotated datasets.

Our work builds upon these advances by integrating LLMs with semantic search techniques to dynamically select the most relevant examples for few-shot learning, thereby enhancing the models' ability to extract events from speech transcriptions.

\section{Dataset}
We use the SpeechEE shared task dataset\footnote{\url{https://xllms.github.io/SpeechEE/}}, derived from ACE2005-EN+. The dataset is provided in a structured JSON format, where each entry consists of a unique \texttt{id}, an event \texttt{trigger} indicating the lexical anchor of the event, its corresponding \texttt{type}, and a list of associated \texttt{arguments}, each annotated with a semantic role. In total, the dataset includes 33 distinct event types and 22 argument roles, offering a diverse and challenging benchmark. It comprises 19,217 training instances, 901 development examples, and 676 test samples. An example of the data format is shown below:

\begin{lstlisting}[basicstyle=\ttfamily\footnotesize, frame=single]
{"id": "train-6", 
 "event": [
   {"trigger": "election", 
    "type": "Elect", 
    "arguments": [
      {"name": "man", "role": "Person"}
    ]}
 ]}
\end{lstlisting}

\section{Methodology}
\begin{figure*}[ht]
\centering
\includegraphics[width=0.95\textwidth]{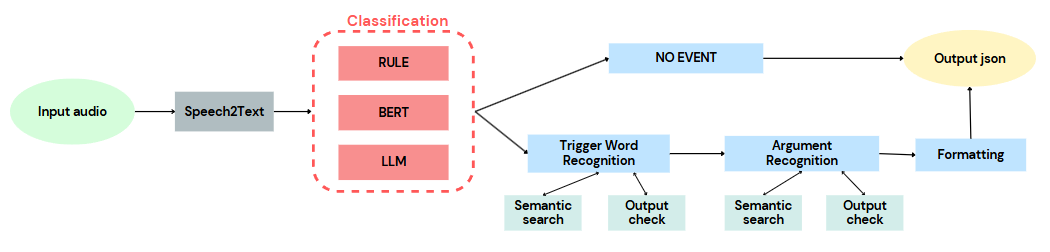}
\caption{Overview of the SpeechEE Pipeline}
\label{fig:pipeline}
\end{figure*}

Our proposed pipeline for speech event extraction consists of several key components, as illustrated in Figure \ref{fig:pipeline}. The pipeline follows a modular approach, allowing for component-level evaluation and optimization.

\subsection{ASR Transcription}
The first step in our pipeline involves transcribing speech data into text. We experimented with two state-of-the-art ASR systems for this purpose, \texttt{Whisper large-v3} and \texttt{Canary 1b}. Both systems were used with their default configurations to transcribe the audio data. The resulting transcripts served as input for subsequent steps in the pipeline.

\subsection{Event Presence Classification}
A significant challenge in event extraction is distinguishing between segments that contain events and those that do not. Preliminary experiments revealed that applying LLMs directly to all transcripts resulted in numerous false positives, where models identified events, that were not annotated in the dataset.

To address this challenge, we implemented a classification step to determine whether a given transcript is likely to contain an event. We employed three different classification methods:

\begin{itemize}
    \item Rule-based approach: This method flagged instances containing trigger words identified in the training set. We compiled a lexicon of trigger words based on the training data and used this to identify potential event-containing segments.
    \item BERT-based classifier: We fine-tuned a BERT model on text embeddings to classify segments as either containing events or not. The model was trained on the transcribed training data with binary labels indicating event presence.
    \item LLM-based classification: We prompted OpenAI's \texttt{o1-mini} model to classify the presence or absence of events in transcripts. The model was given a short description of the task and asked to determine whether a given transcript contained an event.
\end{itemize}

For the BERT-based classification, we used the \texttt{all-MiniLM-L6-v2} sentence transformer model\footnote{\url{https://huggingface.co/sentence-transformers/all-MiniLM-L6-v2}}. The training ran for 5 epochs, with a learning rate of $2e^{-5}$, batch size of 16, choosing the best model based on recall, as the goal was to have as few false negatives as possible. 

To enhance classification reliability, we only considered transcripts in which all three models agreed on the presence of an event. This approach effectively reduced false positives during the classification stage. Table~\ref{tab:agreement} presents the agreement matrix among the models, with bolded cells indicating cases where the filtering mechanism identified likely event presence.

\begin{table}[h]
\centering
\resizebox{0.9\columnwidth}{!}{
\begin{tabular}{llcc}
\toprule
\textbf{Rule} & \textbf{BERT} & \textbf{LLM: NO} & \textbf{LLM: YES} \\
\midrule
\multirow{2}{*}{NO}  & NO   & 258 & 27  \\
     & YES  & 61  & 39  \\
\multirow{2}{*}{YES} & NO   & 17  & 27  \\
     & YES  & 19  & \textbf{228} \\
\bottomrule
\end{tabular}}
\caption{Agreement table between the three systems.}
\label{tab:agreement}
\end{table}

\subsection{Trigger Word Recognition}
For segments classified as likely containing events, the next step involved identifying and classifying trigger words. Trigger words are specific words or phrases that signal the occurrence of an event. These events also have an event type, which needs to be recognized based on the context.

We evaluated three different LLMs for this task: \texttt{Llama3-8B}, \texttt{GPT-4o-mini}, and OpenAI's \texttt{o1-mini}. We deployed \texttt{Llama3–8B} locally on two NVIDIA GeForce RTX 2080 Ti GPUs, while the two OpenAI models were accessed via OpenAI’s Batch API. Each model was prompted to extract trigger words from the transcript and classify them into predefined event categories based on the ACE2005 ontology. The prompt included a description of the task, examples of trigger words for different event types, and the transcript to be analyzed.

During our experiments, we observed that the Llama model occasionally produced outputs that did not comply with the expected format or failed to identify trigger words correctly. To address this issue, we implemented an automated verification step that checked the output format and re-executed queries when necessary to ensure consistency. This step was omissible for the OpenAI models.

\subsection{Semantic Search-Enhanced Few-Shot Learning}
A key component in our approach is the use of semantic search to dynamically select the most relevant examples for few-shot learning. As there are 33 classes, each with multiple argument types, even showing one example from each case would result in a very long prompt. Therefore, rather than using a fixed set of examples for all queries, we implemented a system that selected examples based on their semantic similarity to the current transcript. This retrieval-augmented few-shot approach increases contextual relevance and allows the model to better adapt to domain-specific nuances. By coupling LLMs with semantic retrieval, we aim to improve robustness and generalization in this complex setting.

The process looks like the following:
\begin{enumerate}
    \item We created embeddings for all training examples using a the \texttt{all-MiniLM-L6-v2} sentence transformer model. 
    \item For each new transcript, we generated an embedding using the same model.
    \item We retrieved the top ten most similar examples to the new transcript using the FAISS library \cite{faisslibrary}.
    \item We added these examples to the prompt (Appendix \ref{app_trig}).
\end{enumerate}

This approach ensures that the LLM receives the most relevant and informative examples for each specific transcript, thereby improving its ability to identify and classify trigger words accurately.

\subsection{Argument Extraction}
Following trigger identification, the next step involved extracting event arguments and assigning roles. Event arguments are entities that participate in the event, and their roles define their relationship to the event (e.g., Agent, Entity, Place).

Similar to the trigger recognition phase, we used LLMs to extract arguments and assign roles (Appendix \ref{app_arg}). The prompt for this task included the transcript, the identified trigger word and event type, and examples of argument extraction for similar event types. We again employed semantic search to select the most relevant examples for few-shot learning, focusing on examples similar to the current transcript. Also we provided the dictionary of the possible argument types for each event type in the prompt.

\subsection{Post-Processing}
The final stage of our pipeline involved post-processing the extracted information to ensure uniform output formatting. We used an additional LLM call (Appendix \ref{app_post}) to format the extracted events, triggers, and arguments into a structured JSON format consistent with the dataset specifications. This was necessary, because sometimes the models included the transcripts themselves in the output, or labeled the keys differently.

This step also included validation checks to ensure that the output met the expected schema and that all required fields were present. In cases where the output did not meet these requirements, additional LLM calls were made to correct and complete the information.

\section{Results}

We evaluated our event extraction pipeline using two primary metrics, following the evaluation protocol outlined by \cite{wang2024}:

\begin{itemize}
  \item \textbf{Trigger Classification (TC):} This metric assesses whether both the predicted event type and trigger span match the ground truth exactly. A prediction is considered correct only if both components align perfectly.
  \item \textbf{Argument Classification (AC):} This stricter metric evaluates the correctness of the predicted argument mention, its semantic role, and the associated event type, requiring full agreement across all elements.
\end{itemize}

For both tasks, we report precision (P), recall (R), and F1-score (F1), with the F1-score serving as the primary measure of overall performance. Table~\ref{tab:f1-summary} summarizes the evaluation results across the three LLMs: \texttt{Llama3-8B}, \texttt{GPT-4o-mini}, and \texttt{o1-mini}. The ASR system used is indicated in parentheses—Whisper (W) or Canary (C). For \texttt{Llama3-8B} and \texttt{GPT-4o-mini}, all three stages of the pipeline utilized the respective models. In the case of \texttt{o1-mini}, however, the final formatting step was performed using \texttt{GPT-4o-mini}, as it achieved perfect results for this task, eliminating the need for the more expensive reasoning model.

\begin{table}[!h]
    \centering
    \resizebox{\columnwidth}{!}{
    \begin{tabular}{lcccccc}
        \toprule
        \textbf{Model} & \textbf{TC-R} & \textbf{TC-P} & \textbf{TC-F1} & \textbf{AC-R}  & \textbf{AC-P} & \textbf{AC-F1} \\
        \midrule
        Llama3-8B (W) & 24.1 & 57.6 & 33.9 & 11.2 & 18.3 & 13.9 \\
        GPT-4o-mini (W) & 36.6 & 67.4 & 47.4 & 17.0 & 24.3 & 20.0 \\
        o1-mini (W) & 59.2 & 65.9 & 62.4 & 26.9 & 27.1 & 27.1 \\
        \midrule
        Llama3-8B (C) & 24.5 & 52.8 & 33.5 & 11.5 & 17.1 & 13.7 \\
        GPT-4o-mini (C) & 36.8 & 65.5 & 47.1 & 16.8 & 23.2 & 19.5 \\
        o1-mini (C) & \textbf{60.8} & \textbf{66.0} & \textbf{63.3} & \textbf{28.0} & \textbf{27.6} & \textbf{27.8} \\
        \bottomrule
    \end{tabular}
    }
    \caption{Precision, recall, and F1-scores (\%) for Trigger Classification (TC) and Argument Classification (AC).}
    \label{tab:f1-summary}
\end{table}

The performance hierarchy among the models is consistent and clear: \texttt{o1-mini} substantially outperforms both \texttt{GPT-4o-mini} and \texttt{Llama3-8B} across all evaluation metrics. In TC, \texttt{o1-mini} achieved an F1-score of 63.3\%, surpassing \texttt{GPT-4o-mini} by over 16 percentage points and \texttt{Llama3-8B} by nearly 30 points. Interestingly, while precision scores between \texttt{o1-mini} and \texttt{GPT-4o-mini} were relatively close, the major difference arose from recall, suggesting that \texttt{GPT-4o-mini} adopted a more conservative prediction strategy, prioritizing precision at the expense of coverage.

As anticipated, performance across all models declined on the more demanding AC task. Nevertheless, \texttt{o1-mini} again demonstrated a clear advantage, achieving an AC F1-score of 27.8\%, more than double that of \texttt{Llama3-8B} and significantly ahead of \texttt{GPT-4o-mini}. Moreover, \texttt{o1-mini} maintained a balanced trade-off between precision and recall, whereas the other two models exhibited weaker recall, often failing to capture all valid argument mentions even when precision remained reasonable.

The consistent superiority of \texttt{o1-mini} across both tasks underscores the potential of specialized reasoning models in information extraction, particularly in achieving a more balanced and robust performance across precision and recall.

Regarding the choice of ASR system, switching between Whisper and Canary resulted in only minor differences (within 1\% F1-score). Given the inherent nondeterminism of LLM outputs—even with identical inputs—the observed variations could stem even from stochastic model behavior rather than from the ASR systems themselves. Notably, while \texttt{Llama3-8B} and \texttt{GPT-4o-mini} performed slightly better with Whisper, \texttt{o1-mini} achieved marginally superior results with Canary. A more comprehensive study, involving multiple evaluation runs per prompt, would be required to draw stronger conclusions about ASR influence.

As a point of reference, \cite{wang2024} reported results on the ACE2005-EN+ dataset, achieving an F1-score of 61.1\% for TC and 23.2\% for AC. While our dataset is a modified version of ACE2005-EN+, it is not identical, and thus direct comparisons should be made cautiously. Nevertheless, our pipeline outperforms these baselines, particularly in AC, where we observe a notable improvement. This advancement likely stems from the incorporation of large language models after the transcription step, enabling more semantically coherent interpretations of spoken input instead of bypassing transcription entirely.

\section{Ablation}
To substantiate our initial observation that LLMs tend to produce a considerable number of false positives, we conducted an ablation study evaluating all three models under various classification configurations. The results, presented in Table~\ref{tab:tc_performance} and Table~\ref{tab:ac_performance}, affirm the value of the classification component. We denote \textit{one+} and \textit{two+} to indicate that at least one or two of the three models classified the instance as containing an event, respectively.

\begin{table}[h]
\centering
\resizebox{\columnwidth}{!}{
\begin{tabular}{lccccccc}
\toprule
\textbf{Model} & \textbf{without} & \textbf{Rule} & \textbf{BERT} & \textbf{o1-mini} & \textbf{one+} & \textbf{two+} & \textbf{three} \\
\midrule
\texttt{Llama3-8B} & 27.4 & 32.4 & 32.0 & 32.7 & 30.6 & 33.0 & 33.9 \\
\texttt{4o-mini}   & 43.8 & 47.1 & 45.8 & 45.1 & 44.7 & 46.1 & 47.4 \\
\texttt{o1-mini}   & 58.8 & 60.7 & 61.8 & 60.3 & 59.8 & 62.4 & 63.4 \\
\bottomrule
\end{tabular}}
\caption{F1-scores (\%) on TC under different classification criteria.}
\label{tab:tc_performance}
\end{table}

When applying a single criterion, each model demonstrated optimal performance with a different method: \texttt{o1-mini} performed best with the \textit{BERT-based} classifier, while \texttt{4o-mini} and \texttt{Llama3-8B} yielded higher scores with the \textit{Rule-based} and \textit{LLM-based} classifiers, respectively. Although relying on the \textit{one+} filtering resulted in marginally lower performance than the best individual setups, it still outperformed the baseline without classification. Notably, aggregating predictions with \textit{two+} and \textit{three} led to consistent F1-score improvements over the standalone classifiers.

\begin{table}[h]
\centering
\resizebox{\columnwidth}{!}{
\begin{tabular}{lccccccc}
\toprule
\textbf{Model} & \textbf{without} & \textbf{Rule} & \textbf{BERT} & \textbf{o1-mini} & \textbf{one+} & \textbf{two+} & \textbf{three} \\
\midrule
\texttt{Llama3-8B} & 13.2 & 13.4 & 15.4 & 15.4 & 14.7 & 15.6 & 13.9 \\
\texttt{4o-mini}   & 19.2 & 19.7 & 20.0 & 20.0 & 19.6 & 20.1 & 20.0 \\
\texttt{o1-mini}   & 25.1 & 25.2 & 26.8 & 26.0 & 25.5 & 26.9 & 27.8 \\
\bottomrule
\end{tabular}}
\caption{F1-scores (\%) on AC under different classification criteria.}
\label{tab:ac_performance}
\end{table}

In the AC task, the outcomes are more nuanced, indicating that strategies yielding gains in TC do not always translate to AC. For instance, while \texttt{4o-mini} achieved the best TC performance with the \textit{Rule-based} method (when considering single criterion), it was outperformed by other approaches in AC. Nonetheless, similarly to TC, both \textit{two+} and \textit{three} yielded improvements over the individual classifiers. However, unlike in TC, the \textit{three} criterion did not universally result in the best performance, benefitting only the \texttt{o1-mini} model.

These consistent enhancements across both tasks confirm the effectiveness of the classification step in reducing false positives and improving overall model performance.

\section{Discussion}
Our experimental results demonstrate that a well-designed pipeline approach can achieve performance comparable or even superior to state-of-the-art end-to-end models for speech event extraction. This finding is significant as it challenges the assumption that end-to-end approaches necessarily outperform pipeline methods for complex tasks. Our results suggest that by leveraging advanced LLMs and intelligent example selection strategies, pipeline approaches can mitigate traditional weaknesses such as error propagation.

Moreover, this approach offer several advantages over end-to-end models, including modularity and interpretability, in the sense that the output of each stage can be examined and is humanly interpretable.

The minimal performance difference observed when comparing Whisper and Canary suggests that transcript quality variations within a certain threshold have limited impact on event extraction outcomes. Both Whisper and Canary have their own characteristics, when it comes to the transcription of names or cities, which are crucial in event extraction. The fact that this did not make too much of a difference, is encouraging for real-world applications, where perfect transcription cannot be guaranteed. However, it is important to note that the performance gap might be more pronounced when comparing with lower-quality ASR systems or when processing audio with significant noise, accents, or other challenging characteristics.

The substantial performance differences observed between the three LLMs highlight the critical role of LLM selection in event extraction efficacy. Model size seemed to be a factor, with the larger proprietary models clearly outperforming the locally runnable Llama model. The reasoning capability also largely seems to help the task, but to make these claims more confidently, a more thorough research would be needed across several models.

During development, we also observed that in several cases, the false positives produced by the LLMs were remarkably close to genuine events that should have been annotated. This suggests that, in some instances, the models may have correctly identified events that were inadvertently missed during the original annotation process.

\section{Conclusion}
This research presents an LLM-driven pipeline for speech event extraction that achieves performance comparable to state-of-the-art end-to-end models. Our approach combines ASR-generated transcripts with semantic search-enhanced few-shot learning to create a modular and interpretable framework for identifying events and their arguments from spoken language. By dynamically selecting examples based on semantic similarity to the current input, our approach ensures that the LLM receives the most relevant and informative context for each specific case. This is particularly important for event extraction, where different event types have distinct patterns, trigger words, and argument structures. The effectiveness of this approach suggests that similar techniques could be beneficial for other complex NLP tasks where pattern recognition plays a crucial role and where diverse examples exist in the training data.

The choice of ASR system showed limited impact on extraction performance, suggesting robustness to transcript quality variations within a reasonable range. However, LLM selection plays a critical role in event extraction efficacy, with larger, more capable models achieving significantly better results.

Several promising directions for future research emerge from this work, including hybrid approaches exploring methods that integrate transcript-based and direct audio-based features that could potentially combine the strengths of pipeline and end-to-end approaches. Our approach does not leverage speech-related cues present in the audio, which could potentially enhance performance if incorporated. With prompt engineering, the current prompting strategies could be further refined and may improve LLM performance without requiring additional computational resources. Another area could be exploring methods to reduce computational requirements, such as model distillation or selective component invocation.

This work advances spoken language understanding by demonstrating that modular pipelines can rival end-to-end models through strategic integration of LLMs and retrieval mechanisms. By decoupling transcription from extraction while maintaining cross-component optimization potential, our framework offers a practical pathway for deploying speech event extraction tools.

\section*{Acknowledgments}

We thank the organizers of the SpeechEE shared task for providing the dataset and evaluation framework that enabled this study. 

This paper benefited from the use of generative AI tools in accordance with the ACL Policy on AI Writing Assistance. Specifically, large language models were used for assistance in paraphrasing and polishing the original content. All such generated content was carefully reviewed and revised by the authors, and relevant citations were added where appropriate. The ideas and scientific contributions presented in this paper are entirely the authors own. The authors bear full responsibility for the accuracy and originality of the work.

\section*{Limitations}
First, although the classification step successfully reduces false positives, it introduces an additional layer of complexity and latency into the pipeline. In real-time applications, this could pose practical constraints. 

Second, although Whisper and Canary represent state-of-the-art multilingual ASR models, their performance varies considerably across languages, limiting generalizability in multilingual settings. This challenge is further compounded by the few-shot prompting technique, which is also expected to yield lower performance for low-resource languages due to limited linguistic and contextual coverage in training data.

Third, although semantic search-enhanced few-shot prompting improves LLM performance, it increases computational costs due to the need for embedding comparisons and dynamic prompt construction. This makes the system more resource-intensive, especially for large-scale or low-latency deployments.

Fourth, the argument classification task remains challenging, with overall performance still low despite LLM assistance. This is likely due to the difficulty of correctly identifying multiple, diverse argument roles within spoken inputs. Improving argument role assignment—especially for less frequent event types—requires further attention, potentially through the use of more structured reasoning or task-specific tuning.

Finally, while the \texttt{o1-mini} model consistently outperformed the other models in our experiments, it is a proprietary model, limiting reproducibility and potentially raising cost or accessibility concerns. Our pipeline's dependency on API-based models also poses challenges for deployment in privacy-sensitive or resource-constrained environments.

Future work should explore more efficient alternatives for few-shot prompting, more robust handling of ASR errors, and ways to make the pipeline more lightweight and adaptable to diverse real-world settings.

\bibliography{main}

\appendix

\clearpage
\onecolumn
\section{Appendix}
\label{sec:appendix}

\subsection{Trigger Recognition Prompt} \label{app_trig}
\begin{tcolorbox}
\small
\begin{verbatim}
{"role": "system", "content": 
 "Your job is to extract trigger words signaling events in a text, and classify its event type."},
{"role": "user", "content": 
 "From the following TEXT, please extract the event type and its trigger word. It is a transcript of 
 an audio, so there may be some mistakes. 
 The possible event types are: [<event type list>]. 
 It is possible there are no events in the text. 
 Below are examples demonstrating the required output format and some useful hints. 
 Do not return the transcript, only the trigger word and event type."},
{"role": "user", "content": "TEXT: <text_input>"},
{"role": "user", "content": "EXAMPLES: <few-shot examples>"},
\end{verbatim}
\end{tcolorbox}

\subsection{Argument Recognition Prompt} \label{app_arg}
\begin{tcolorbox}
\small
\begin{verbatim}
{"role": "system", "content": 
 "Your job is to extract arguments for events in a text, and classify their role in that event."},
{"role": "user", "content": 
 "From the following TEXT, please extract event arguments (usually one word or a name) and their role. 
 It is a transcript of audio, so there may be mistakes. 
 Use the provided event schema: <event schema>. 
 An event may have no arguments. 
 Examples are provided to guide selection and format."},
{"role": "user", "content": "TEXT: <text_input>, EVENT TYPE(s): <event types>"},
{"role": "user", "content": "EXAMPLES: <few-shot examples>"},
\end{verbatim}
\end{tcolorbox}

\subsection{Post-processing Prompt} \label{app_post}
\begin{tcolorbox}
\small
\begin{verbatim}
{"role": "system", "content": 
 "Your job is to extract a JSON-like output from the end of a string. Only return the JSON."},
{"role": "user", "content": 
 "From the following TEXT, extract data in the format of the example. 
 If multiple triggers exist, return one entry per trigger. 
 Transcriptions are unnecessary. Return JSON only."},
{"role": "user", "content": "TEXT: <text_input>"},
{"role": "user", "content": 
 "EXAMPLE: 
  [
    {
      "trigger": "deploy",
      "type": "Transport",
      "arguments": [
        {"name": "soldiers", "role": "Artifact"},
        {"name": "region", "role": "Destination"}
      ]
    }
  ]"},
\end{verbatim}
\end{tcolorbox}

\end{document}